\documentclass{article}

\PassOptionsToPackage{numbers}{natbib}

\usepackage[preprint]{neurips_2020}

\usepackage{neurips_2020}
\usepackage[T1]{fontenc}    
\usepackage{hyperref}       
\usepackage{url}            
\usepackage{booktabs}       
\usepackage{amsfonts}       
\usepackage{nicefrac}       
\usepackage{microtype}      

\usepackage{bbm}
\usepackage{color}
\usepackage{wrapfig}        
\usepackage{graphicx}
\usepackage{subcaption}
\usepackage{algorithm}
\usepackage{algorithmic}
\usepackage{amsmath,amsfonts,bm}
\usepackage{amssymb}
\usepackage{enumitem}
\usepackage{makecell}
\usepackage{multirow}

\usepackage{amsmath,amsfonts,bm}

\def\vp{{\bm{p}}}

\def\vx{{\bm{x}}}

\DeclareMathAlphabet{\mathsfit}{\encodingdefault}{\sfdefault}{m}{sl}
\SetMathAlphabet{\mathsfit}{bold}{\encodingdefault}{\sfdefault}{bx}{n}

\title{A Universal Representation Transformer Layer\\ for Few-Shot Image Classification}

\author{%
  Lu Liu$^{1,2}$\thanks{This work was done while Lu Liu was a research intern with Mila.}, William Hamilton$^{1,3}$\thanks{Canada CIFAR AI Chair} , Guodong Long$^{2}$, Jing Jiang$^{2}$, Hugo Larochelle$^{1,4\dagger}$ \\
$^{1}$ Mila, $^{2}$ Australian AI Institute, UTS, $^{3}$ McGill University,  $^{4}$ Google Research, Brain Team\\
  Correspondence to \texttt{\href{mailto:lu.liu.cs@icloud.com}{lu.liu.cs@icloud.com}}
}

\newcommand*\rot{\rotatebox{0}}

\begin{document}

\maketitle

\begin{abstract}
Few-shot classification aims to recognize unseen classes when presented with only a small number of samples. 
We consider the problem of {\em multi-domain} few-shot image classification, where unseen classes and examples come from diverse data sources. This problem has seen growing interest and has inspired the development of benchmarks such as Meta-Dataset. A key challenge in this multi-domain setting is to effectively integrate the feature representations from the diverse set of training domains.
Here, we propose a Universal Representation Transformer (URT) layer, that meta-learns to leverage universal features for few-shot classification by dynamically re-weighting and composing the most appropriate domain-specific representations.
In experiments, we show that URT sets a new state-of-the-art result on Meta-Dataset. 
Specifically, it achieves top-performance on the highest number of data sources compared to competing methods.
We analyze variants of URT and present a visualization of the attention score heatmaps that sheds light on how the model performs cross-domain generalization. Our code is available at \texttt{\href{https://github.com/liulu112601/URT}{https://github.com/liulu112601/URT}}
\end{abstract}

\section{Introduction}

Learning tasks from small data remains a challenge for machine learning systems, which show a noticeable gap compared to the ability of humans to understand new concepts from few examples. A promising direction to address this challenge is developing methods that are capable of performing transfer learning across the collective data of many tasks. 
Since machine learning systems generally improve with the availability of more data, a natural assumption is that few-shot learning systems should benefit from leveraging data {\em across many different tasks and domains}---even if each individual task has limited training data available.

This research direction is well captured by the problem of {\em multi-domain few-shot classification}. 
In this setting, training and test data spans a number of different domains, each represented by a different source dataset. 
A successful approach in this multi-domain setting must not only address the regular challenge of few-shot classification---i.e., the challenge of having only a handful of examples per class. 
It must also discover how to leverage (or ignore) what is learned from different domains, achieving generalization and avoiding cross-domain interference.

Recently, \citet{triantafillou2019meta} proposed a benchmark for multi-domain few-shot classification, Meta-Dataset, and highlighted some of the challenges that current methods face when training data is heterogeneous. Crucially, they found that methods which trained on all available domains would normally obtain improved performance on some domains at the expense of others. Following on their work, progress has been made, which includes the design of adapted hyper-parameter optimization strategies~\cite{saikia2020optimized} and more flexible meta-learning algorithms~\cite{requeima2019fast}. Most notable is SUR (Selecting Universal Representation)~\cite{dvornik2020selecting}, a method that relies on a so-called universal representation, extracting from a collection of pre-trained and domain-specific neural network backbones. SUR prescribes a hand-crafted feature-selection procedure to infer how to weight each backbone for each task at hand, and produces an adapted representation for each task. This was shown to lead to some of the best performances on Meta-Dataset.

In SUR, the classification procedure for each task is fixed and not learned. Thus, except for the underlying universal representation, there is no transfer learning performed with regards to how classification rules are inferred across tasks and domains. Yet, cross-domain generalization might be beneficial in that area as well, in particular when tasks have only few examples per class. 

\textbf{Present work.}
To explore this question, we propose the Universal Representation Transformer (URT) layer, which can effectively learn to transform a universal representation into task-adapted representations. 
The URT layer is inspired from Transformer networks~\cite{vaswani2017attention} and uses an attention mechanism to learn to retrieve or blend the appropriate backbones to use for each task. By training this layer across few-shot tasks from many domains, it can support transfer across these tasks.

We show that our URT layer on top of a universal representation's pre-trained backbones sets a new state-of-the-art performance on Meta-Dataset.
It succeeds at outperforming SUR on 4 dataset sources without impairing accuracy on the others. This leads to top performance on 7 dataset sources when comparing to a set of competing methods.
To interpret the strategy that URT learns to weigh the backbones from different domains, we visualize the attention scores for both seen and unseen domains and find that our model generates meaningful weights for the pre-trained domains. A comprehensive analysis on variants and ablations of the URT layer is provided to show the importance of various components of URT, notably the number of attention heads.

\section{Few-Shot Classification}

\subsection{Problem Setting} 
In this section, we will introduce the problem setting for few-shot classification and the formulation of meta-learning for few-shot classification. Few-shot classification aims to classify samples where only few examples are available for each class. We describe a few-shot learning classification task
as the pair of examples, comprising of a support set $S$ to define the classification task and the query set $Q$ of samples to be classified.

Meta-learning is a technique that aims to model the problem of few-shot classification as learning to learn from instances of few-shot classification tasks. The most popular way to train a meta-learning model is with episodic training. Here, tasks $T = (Q, S)$ are sampled from a larger dataset by taking subsets of the dataset to build a support set $S$ and a query set $Q$ for the task. A common approach is to sample $N$-way-$K$-shot tasks, each time selecting a random subset of $N$ classes from the original dataset and choosing only $K$ examples for each class to add to the support set $S$.

The meta-learning problem can then be formulated by the following optimization:
\begin{align}\label{equ:opt-obj}
\min_\Theta\mathbb{E}_{(S,Q)\sim p(T)}\left[{\cal L}(S,Q,\Theta)\right],~{\cal L}(S,Q,\Theta)=\frac{1}{|Q|}\sum_{(\vx,y)\sim Q}  -\log p(y|\vx, S; \Theta) + \lambda \Omega(\Theta),
\end{align}
where $p(T)$ is the distribution of tasks, $\Theta$ are the parameters of the model and $p(y|\vx, S; \Theta)$ is the probability assigned by the model to label $y$ of query example $\vx$ (given the support set $S$), and $\Omega(\Theta)$ is an optional regularization term on the model parameters with factor $\lambda$.

Conventional few-shot classification targets the setting of $N$-way-$K$-shot, where the number of classes and examples are fixed in each episode. Popular benchmarks that follow this approach include Omniglot~\cite{lake2015human}) or benchmarks made of subsets of ImageNet, such as \textit{mini}ImageNet~\cite{vinyals2016matching} and \textit{tiered}ImageNet~\cite{ren2018meta}. In such benchmarks, the tasks used for training cover a set of classes that is disjoint from the classes found in the test set of tasks. However, with the training and test sets tasks coming from a single dataset/domain, the distribution of tasks found in either sets is similar and lacks variability, which may be unrealistic in practice.

It is in this context that \citet{triantafillou2019meta} proposed Meta-Dataset, as a further step towards large-scale, multi-domain few shot classification. 
Meta-Dataset includes ten datasets (domains), with eight of them available for training. Additionally, each task sampled in the benchmark varies in the number of classes $N$, with each class also varying in the number of shots $K$. As in all few-shot learning benchmarks, the classes used for training and testing do not overlap.

\subsection{Background and Related Work} 

{\bf Transfer by fine-tuning} 
A simple and effective method for few-shot classification is to perform transfer learning by first learning a neural network classifier on all data available for training and using its representation to initialize and then fine-tune neural networks on the few-shot classification tasks found at test time~\cite{closerlook,triantafillou2019meta,DhillonG2020,saikia2020optimized}. Specifically, \citet{saikia2020optimized} have shown that competitive performance can be reached using a strong hyper-parameter optimization method applied on a carefully designed validation metric appropriate for few-shot learning.

{\bf Meta-Learning} Another approach is to use meta-learning to more directly train a model to learn to perform few-shot classification, in an end-to-end way.
A large variety of such models have been explored, inspired by
memory-augmented networks~\cite{santoro2016meta}, LSTMs~\cite{ravi2016optimization} and metric-based classifiers~\cite{vinyals2016matching}.
The two most popular methods are Prototypical Networks~\cite{snell2017prototypical} and Model Agnostic Meta-Learning (MAML)~\cite{finn2017model}. Prototypical Networks assume that every class can be represented as a prototype in a learned embedding space (represented by a neural network). Prototypes are calculated as the average of the representations of the support examples of each class. A 1-nearest centroid classifier is then adopted for classification and the neural network representation is trained to facilitate classification in few-shot tasks directly.
MAML models the procedure of learning to learn as a bilevel optimization, where an outer loop backpropagates loss gradients through an optimization-based inner loop to learn its initialization.
\citet{triantafillou2019meta} showed that prototypical networks and MAML could be combined by leveraging prototypes for the initialization of the output weights value in the inner loop. 
\citet{requeima2019fast} also proposed Conditional Neural Adaptive Processes (CNAPs) for few-shot classification, which can be seen as extending prototypical networks with a more sophisticated architecture that allows for improved task adaptation. This architecture was later improved further by \citet{bateni2020improved} with Simple~CNAPS, leading to one of the current best methods on Meta-Dataset.
\looseness-1

{\bf Universal Representatons} In contrast, our work instead builds on that of \citet{dvornik2020selecting} and their method SUR (Selecting from Universal Representations). \citet{BilenH2017} introduced the term {\it universal representation} to refer to a representation that supports good performance in multiple domains. One proposal towards such a representation is to train different neural networks backbones separately on the data of each available domain, then simply to concatenate the representation learned by each. Another is to introduce some parameter sharing between the backbones, by having a single network conditioned on the domain of the provenance of each batch of training data~\cite{RebuffiS2018}, e.g.\ using Feature-wise Linear Modulate (FiLM)~\cite{film2018}.
SUR proposes to leverage a universal representation in few-shot learning tasks with a feature selection procedure that assigns different weights to each of the domain-specific subvectors of the universal representation. The objective is to assign high weights only to the domain-specific representations that are specifically useful for each few-shot task at hand. The weights are inferred by optimizing a loss  on the support set that encourages high accuracy of a nearest-centroid classifier. As such, the method does not involve any meta-learning---a choice motivated by the concern that  meta-learning may struggle in generalizing to domains that are dissimilar to the training domains. 
SUR achieved some of the best performances on Meta-Dataset. However, a contribution of our work is to provide evidence that meta-learning can actually be used to replace SUR's hand-designed inference procedure and improve performance further.

{\bf Transformer Networks} Our meta-learning approach to leverage universal representations is inspired directly from Transformer networks~\cite{vaswani2017attention}. Transformer networks are neural network architectures characterized by the use self-attention mechanisms to solve tasks. Our model structure is inspired by the structure of the dot-product self-attention in the Transformer, which we adapted here to multi-domain few-shot learning by designing appropriate parametrizations for queries, keys and values. Self-attention was explored in the single-domain training regime by \citet{YeH2020}, however for a different purpose, where each representation of individual examples in a task support set is influenced by all other examples. Such a strategy is also employed by CNAPs, but with the latter using FiLM as the conditioning mechanism, instead of self-attention.
Regardless, the aim of this paper is to propose a different strategy.  
Rather than using self-attention between individual examples in the support set, our model uses self-attention to select between different domain-specific backbones.

\begin{figure}[t!]
\begin{center}
\includegraphics[width=\linewidth]{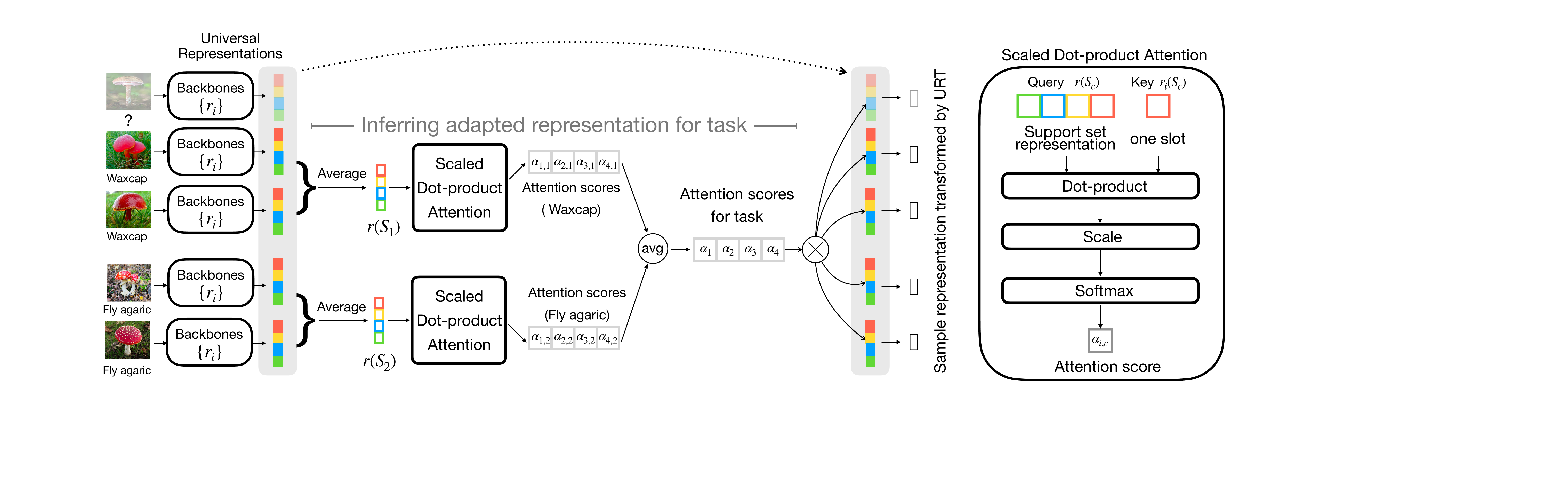}
\end{center}
\caption{Illustration of how a single-head URT layer uses a universal representation to produce a task-specific representation. This example assumes the use of four backbones, with each color illustrating their domain-specific sub-vector representation in the universal representation. 
}
\label{fig:urt}
\end{figure}

\section{Universal Representation Transformer Layer}
\label{sec:urt-layer}

In this section, we describe our proposed URT layer, which uses meta-learning episodic training to learn how to combine the domain-specific backbones of a universal representation for any given few-shot learning classification task.

Conceptually, the proposed model views the support set $S$ of a task as providing information on how to query and retrieve from the set $\{r_i\}$ of $m$ pre-trained backbones the most appropriate backbone to build an adapted representation $\phi$ for the task. 

We would like the model to support a variety of strategies on how to retrieve backbones. For example, it might be beneficial for the model to retrieve a single backbone from the set, especially if the domain of the given task matches perfectly that of a domain found in the training set. Alternatively, if some of the training domains benefit from much more training data than others, a better strategy might be to attempt some cross-domain generalization towards the few-shot learning task by blending many backbones together, even if none matches
the domain of the task perfectly.

This motivates us to use dot-product self-attention, inspired by layers of Transformer networks~\cite{vaswani2017attention}. For this reason, we refer to our model as a Universal Representation Transformer (URT) layer.
Additionally, since each class of the support set might require a different strategy, we perform attention separately for each class and their support set $S_c = \{\vx | (\vx,y) \in S~{\rm and}~y=c\}$.

\subsection{Single-Head URT Layer}

We start by describing an URT layer consisting of a single attention head. An illustration of a single-head URT layer is shown in Figure~\ref{fig:urt}.
Let $r_i(\vx)$ be the output vector of the backbone for domain $i$. We then write the universal representation as
\begin{align}
\label{eq:sample-rep}
    r({\bf x} ) = {\rm concat}(  r_1({\bf x}),\dots,  r_m(\bf x)) .
\end{align}
This representation provides a natural starting point to obtain a representation of a support set class. Specifically, we will note
\begin{align}
\label{eq:set-rep}
    r(S_c) = \frac{1}{|S_c|} \sum_{\vx \in S_c} r(\vx)
\end{align}
as the representation for the set $S_c$.
From this, we can describe the URT layer by defining the queries\footnote{Unable to avoid the unfortunate double usage of the term "query" due to conflicting conventions, we highlight the difference between the {\it query} sets $Q$ of few-shot tasks and the {\it queries} ${\bf q}_c$ of an attention mechanism.}, keys, the attention mechanism and output of the layer:

{\bf Queries ${\bf q}_c$:} For each class $c$, we obtain a query through ${\bf q}_c = {\bf W}^q  r(S_{c}) + {\bf b}^q$, where we have a learnable query linear transformation represented by matrix ${\bf W}^q$ and bias ${\bf b}^q$.
    
{\bf Keys ${\bf k}_{i,c}$:} For each domain $i$ and class $c$, we define keys as ${\bf k}_{i,c} = {\bf W}^k  r_{i}(S_{c}) + {\bf b}^k$, using a learnable linear transformation ${\bf W}^k$ and ${\bf b}^k$ and where $r_{i}(S_{c}) = 1/|S_c| \sum_{\vx \in S_c} r_i(\vx) $, using a similar notation as for $r(S_c)$.
    
{\bf Attention scores $\alpha_i$:} as for regular Transformer layers, we use scaled dot-product attention
\begin{align}
\label{eq:class-attention}
 \alpha_{i,c} = \frac{\exp(\beta_{i,c})}{\sum_{i'} \exp(\beta_{i',c})},
 \beta_{i,c} = \frac{{{\bf q}_c}^\top {\bf k}_{i,c}}{\sqrt{l}}, 
\end{align}
where $l$ is the dimensionality of the keys and queries. Then, these per-class scores are aggregated to obtain scores for the full support set by averaging
\begin{align}
\label{eq:urt-attention}
    \alpha_{i} = \frac{\sum_c \alpha_{i,c}}{N}.
\end{align}

Equipped with these attention scores, the URT layer can now produce an adapted representation for the task (for the support and query set examples) by computing
\begin{align}
\label{eq:urt-layer}
    \phi({\bf x}) =  \sum_i \alpha_{i} r_{i}({\bf x})~.
\end{align}
As we can see, this approach has the flexibility of either selecting a single domain-specific backbone (by assigning $\alpha_i = 1$ for a single domain) or blending different domains together (by having $\alpha_i >> 0$ for multiple backbones).

\subsection{Multi-Head URT Layer}
The URT layer described so far can only learn to retrieve a single backbone (or blending of backbones). Yet, it might be beneficial to retrieve multiple different (blended) backbones, especially for a few-shot task that would include many classes of varying complexity.

Thus, to achieve such diversity in the adapted representation,  we also consider URT layers with multiple heads, i.e.\ where each head corresponds to the calculation of Equation~\ref{eq:urt-layer} and each head has its own set of parameters (${\bf W}^q, {\bf b}^q, {\bf W}^k, {\bf b}^k$). Denoting each head now as $\phi_h$, a multi-head URT layer then produces as its output the concatenation of all of its heads:
\begin{align}
\label{eq:multi-head-urt}
   \phi({\bf x}) = \rm concat(\phi_1(\vx), \dots, \phi_H(\vx)).
\end{align}
Empirically we found that the randomness in the initialization of head weights alone did not lead to uniqueness and being complimentary between the heads, so inspired by~\citet{lin2017structured}, we add a regularizer to avoid duplication of the attention scores:
\begin{align}
\label{eq:regularizer}
    \Omega(\Theta) = {\| ( {\bf AA^\top - I} )\|_{F}}^{2},
\end{align}
where $\|\cdot\|_{F}$ is the Frobenius norm of a matrix and ${\bf A} \in \mathbb{R}^{n \times m}$ is the matrix for attention scores, with ${\bf A}_h$ being the vector of all scores $\alpha_i$ for head $h$. The identity matrix $\bf I$ regularizes each set of attention scores to be more focused so that multiple heads can attend to different domain-specific backbones.

\begin{algorithm}[t]
\caption{Training of URT layer }
\label{alg:urt}
\begin{algorithmic}[1]
\REQUIRE  Number of tasks $\tau_{total}$, $m$ pre-trained backbones ;
\FOR{$\tau\in\{1,\cdots,\tau_{total}\}$}
\STATE Sample a few-shot task $T$ with support set $S$ and query set $Q$;
\STATE {\bf \# Infer adapted representation for task from $S$}
\STATE For each class, obtain representation using $m$ pre-trained backbones as in Eq.~(\ref{eq:set-rep});
\STATE Obtain attention scores using Eq.~(\ref{eq:class-attention},\ref{eq:urt-attention}) for each head using support set $S$;
\STATE {\bf \# Use adapted representation to predict labels in $Q$ from support set $S$}
\STATE Compute adapted representation of examples in $S$ and $Q$ as in Eq.~(\ref{eq:urt-layer},\ref{eq:multi-head-urt});
\STATE Compute probabilities of label of examples in $Q$ using Prototypical Network as in Eq.~(\ref{equ:classification});
\STATE Compute loss as in Eq.~(\ref{equ:opt-obj},\ref{eq:regularizer}) and perform gradient descent step on URT parameters $\Theta$;
\ENDFOR
\end{algorithmic}
\end{algorithm}

\subsection{Training Strategy}
We train representations produced by the URT layer by following the approach of Prototypical Networks~\cite{snell2017prototypical}, where the probability of a label $y$ for a query example $\vx$ given the support set of a task is modeled as:
\begin{align}\label{equ:classification}
p(y=c|\vx, S; \Theta) = \frac{\exp( -d( \phi(\vx) - \vp_c) )}{\sum_{c'=1}^N \exp( -d( \phi(\vx) -\vp_{c'}))},
\end{align}
where $d$ is a distance metric and $\vp_c = 1/|S_c| \sum_{\vx \in S_c} \phi(\vx)$ corresponds to the centroid of class $c$, referred to as its prototype. We use (negative) cosine similarity as the distance.
The full training algorithm is presented in Algorithm~\ref{alg:urt}.

\section{Experiments}

In this section, we seek to answer three key experimental questions:
\begin{enumerate}[label=\textbf{Q\arabic*}, itemsep=2pt, parsep=2pt, leftmargin=*, topsep=2pt]
    \item How does URT compare with previous state-of-the-art on Meta-Dataset for multi-domain few-shot classification?
    \item Do the URT attention heads generate interpretable and meaningful attention scores? 
     \item Does the URT layer provide consistent benefits, even when pre-trained backbones are trained in different ways? 
\end{enumerate}
In addition, we investigate architectural choices made, such as our models for keys/queries and their regularization, and study their contribution to achieving strong performance with URT.

\begin{table}[t!]
\scriptsize
\centering
\setlength{\tabcolsep}{2.0pt}
\caption{
Test performance (mean+CI\%95) over 600 few-shot tasks. URT and the most recent methods, which are listed in the first row, are compared on Meta-Dataset~\cite{triantafillou2019meta}, which are listed in the first row. The numbers in \textbf{bold} have intersecting confidence intervals with the most accurate method. 
}
\vspace{0.5em}
\begin{tabular}{lcccccccc|cc|c}
\toprule
& \textbf{\rotatebox{45}{ILSVRC}} & \textbf{\rotatebox{45}{Omniglot}} & \textbf{\rotatebox{45}{Aircraft}} & \textbf{\rotatebox{45}{Birds}} & \textbf{\rotatebox{45}{Textures}} & \textbf{\rotatebox{45}{QuickDraw}} & \textbf{\rotatebox{45}{Fungi}} & \textbf{\rotatebox{45}{VGGFlower}} & \textbf{\rotatebox{45}{TrafficSigns}} & \textbf{\rotatebox{45}{MSCOCO}} & \textbf{\rotatebox{45}{avg. rank}}  \\
\midrule
\textbf{MAML}\cite{finn2017model} & 37.8$\pm$1.0 &  83.9$\pm$1.0 &  76.4$\pm$0.7 &  62.4$\pm$1.1 &  64.1$\pm$0.8 &  59.7$\pm$1.1 &  33.5$\pm$1.1 &  79.9$\pm$0.8 &  42.9$\pm$1.3 &  29.4$\pm$1.1 & 8.0 \\
\textbf{ProtoNet}\cite{snell2017prototypical} & 44.5$\pm$1.1 & 79.6$\pm$1.1 & 71.1$\pm$0.9 & 67.0$\pm$1.0 & 65.2$\pm$0.8 & 64.9$\pm$0.9 & 40.3$\pm$1.1 & 86.9$\pm$0.7 & 46.5$\pm$1.0 & 39.9$\pm$1.1 & 7.3 \\
\textbf{ProtoMAML}\cite{triantafillou2019meta} & 46.5$\pm$1.1 &  82.7$\pm$1.0 &  75.2$\pm$0.8 &  69.9$\pm$1.0 &  68.3$\pm$0.8 &  66.8$\pm$0.9 &  42.0$\pm$1.2 &  88.7$\pm$0.7 &  52.4$\pm$1.1 &  41.7$\pm$1.1 & 5.4  \\
\textbf{CNAPs}\cite{requeima2019fast} & 52.3$\pm$1.0 &  88.4$\pm$0.7 &  80.5$\pm$0.6 & 72.2$\pm$0.9 & 58.3$\pm$0.7 & 72.5$\pm$0.8 & 47.4$\pm$1.0 & 86.0$\pm$0.5 & 60.2$\pm$0.9 & 42.6$\pm$1.1 & 5.1 \\
\textbf{BOHB-E}\cite{saikia2020optimized} & 55.4$\pm$1.1 & 77.5$\pm$1.1 & 60.9$\pm$0.9 &   73.6$\pm$0.8 & \textbf{72.8$\pm$0.7} &  61.2$\pm$0.9  &  44.5$\pm$1.1 & \textbf{90.6$\pm$0.6} & 57.5$\pm$1.0 & \textbf{51.9$\pm$1.0} & 4.4 \\
\textbf{TaskNorm}\cite{bronskill2020tasknorm} & 50.6$\pm$1.1 & 90.7$\pm$0.6 & 83.8$\pm$0.6 & 74.6$\pm$0.8 & 62.1$\pm$0.7 & 74.8$\pm$0.7 & 48.7$\pm$1.0 & 89.6$\pm$0.6 & 67.0$\pm$0.7 & 43.4$\pm$1.0 & 3.8 \\
\textbf{SUR}\cite{dvornik2020selecting} & 56.3$\pm$1.1 & 93.1$\pm$0.5  &  \textbf{85.4$\pm$0.7} &  71.4$\pm$1.0 &  \textbf{71.5$\pm$0.8} &  81.3$\pm$0.6 &  \textbf{63.1$\pm$1.0} &  82.8$\pm$0.7 & 70.4$\pm$0.8 & \textbf{52.4$\pm$1.1} & 2.5 \\
\textbf{SimpleCNAPS}\cite{bateni2020improved} & \textbf{58.6$\pm$1.1} & 91.7$\pm$0.6 & 82.4$\pm$0.7 & 74.9$\pm$0.8 & 67.8$\pm$0.8 & 77.7$\pm$0.7 & 46.9$\pm$1.0 & \textbf{90.7$\pm$0.5} & \textbf{73.5$\pm$0.7} & 46.2$\pm$1.1 & 2.4 \\
\textbf{URT} & 55.7$\pm$1.0 &  \textbf{94.4$\pm$0.4} &  \textbf{85.8$\pm$0.6} & \textbf{76.3$\pm$0.8} & \textbf{71.8$\pm$0.7} & \textbf{82.5$\pm$0.6} & \textbf{63.5$\pm$1.0} & 88.2$\pm$0.6 & 69.4$\pm$0.8 & \textbf{52.2$\pm$1.1} & 1.6 \\

\bottomrule
\end{tabular}
\label{table:main-results}
\end{table}

\subsection{Datasets and Setup}
We test our methods on the large-scale few-shot learning benchmark Meta-Dataset~\cite{triantafillou2019meta}. It consists of ten datasets with various data distributions across different domains, including natural images (Birds, Fungi, VGG Flower), hand-written characters (Omniglot, Quick Draw), and human created objects (Traffic Signs, Aircraft). Among the ten datasets, eight provide data that can be used during either training, validation and testing (with each class assigned to only one of those sets), while two datasets are solely used for testing. Following~\citet{bateni2020improved,requeima2019fast}, we also report results on MNIST~\cite{lecun1998gradient}, CIFAR10 and CIFAR100~\cite{krizhevsky2009learning} as additional unseen test datasets. Following \citet{triantafillou2019meta}, few-shot tasks are sampled with varying number of classes $N$, varying number of shots $K$ and class imbalance. 
The performance is reported as the average accuracy over 600 sampled tasks. More details of Meta-Dataset can be found in~\citet{triantafillou2019meta}.

The domain-specific backbones are pre-trained following the setup in~\cite{dvornik2020selecting}. Then, we freeze the backbone and train the URT layer for 10,000 episodes, with an initial learning rate of 0.01 and a cosine learning rate scheduler. 
Following~\citet{chen2020new}, the training episodes have 50\% probability coming from the ImageNet data source.
Since different pre-trained backbones may produce representations with different vector norms, we normalize the outputs of the backbones as in \citet{dvornik2020selecting}.
URT is trained with parameter weight decay of 1e-5 and with a regularization factor $\lambda = 0.1$. The number of heads ($H$ in Equation~\ref{eq:multi-head-urt}), is set to 2 and the dimension of the keys and queries ($l$ in Equation~\ref{eq:class-attention}) is set to 1024.
We choose the hyper-parameters based on the performance of the validation set. Details of the hyper-parameter selection and how the performance is influenced by them are outlined in Section~\ref{sec:hyper-parameter-selection}.

\subsection{Comparison with Previous Approaches}
\label{ref:main-result}
Table~\ref{table:main-results} presents a comparison of URT with SUR, as well as other baselines based on transfer learning by fine-tuning~\cite{saikia2020optimized} or meta-learning (Prototypical Networks~\cite{snell2017prototypical}, first-order MAML~\cite{finn2017model}, ProtoMAML~\cite{triantafillou2019meta}, CNAPs~\cite{requeima2019fast}) and Simple~CNAPS\cite{bateni2020improved}.

\begin{wraptable}{r}{0.7\textwidth}
\centering
\setlength{\tabcolsep}{2.5pt}
\caption{
Test performance (mean+CI\%95) over 600 few-shot tasks on additional datasets.
}
\vspace{0.5em}
\begin{tabular}{lcccc}
\toprule
& \textbf{MNIST} & \textbf{CIFAR10} & \textbf{CIFAR100} & \textbf{avg. rank} \\
\midrule
\textbf{CNAPs}\cite{requeima2019fast} & 92.7 $\pm$ 0.4 & 61.5 $\pm$ 0.7 & 50.1 $\pm$ 1.0 & 4.7 \\
\textbf{TaskNorm}\cite{bronskill2020tasknorm} & 92.3 $\pm$ 0.4 & 69.3 $\pm$ 0.8 & 54.6 $\pm$ 1.1 & 3.3 \\
\textbf{SUR}\cite{dvornik2020selecting} & {\bf94.3 $\pm$ 0.4} & 66.8 $\pm$ 0.9 & 56.6 $\pm$ 1.0 & 2.3  \\
\textbf{SimpleCNAPS}\cite{bateni2020improved} & 93.9 $\pm$ 0.4 & \textbf{74.3 $\pm$ 0.7} & \textbf{60.5 $\pm$ 1.0} & 1.7 \\
\textbf{URT} & {\textbf{94.8 $\pm$ 0.4}} & 67.3 $\pm$ 0.8 & 56.9 $\pm$ 1.0 & 2.0\\

\bottomrule
\end{tabular}
\label{table:additional-results}
\end{wraptable}

We observe in Table~\ref{table:main-results} that URT establishes a new state-of-the-art on Meta-Dataset, by achieving the top performance on 7 out of the 10 dataset sources. When comparing to its predecessor, URT outperforms SUR on 4 datasets without compromising performance on others, which is challenging to achieve in the multi-domain setting.
Of note, the average inference time for URT is 0.04 second per task, compared to 0.43 for SUR, on a single V100. Thus, getting rid of the optimization procedure for every episode with our meta-trained URT layer also significantly increases the latency, by more than 10$\times$.

We also report performances on the MNIST, CIFAR-10 and CIFAR-100 dataset sources in Table~\ref{table:additional-results}, and compare with the subset of methods that have reported on these datasets. There, URT neither improves nor gets worse performance than SUR, yeilding top performance on the MNIST domain but not on the CIFAR-10/CIFAR-100 domain, on which Simple~CNAPS has the best performance.

\begin{figure}[t!]
\begin{center}
\includegraphics[width=\linewidth]{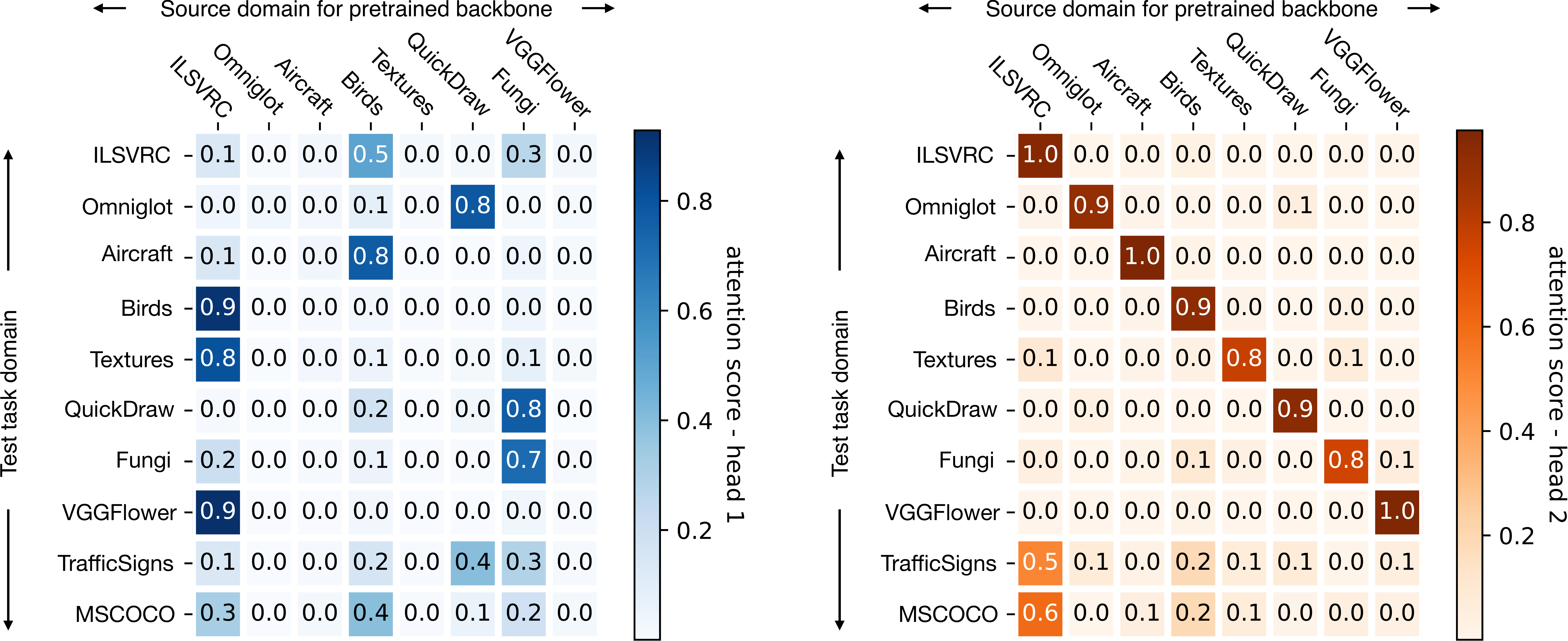}
\end{center}
\caption{
Average attention scores generated by URT with two heads. 
Rows correspond to the domain of the test tasks and the columns correspond to the pre-trained backbones $r_i(\vx)$ trained on the eight training domains.}
\label{fig:attention-score}
\end{figure}

\subsection{Interpreting and Visualizing Attention by URT}
To better understand how the URT model of Section~\ref{ref:main-result} uses its two heads to build adapted representations, we visualize the attention scores produced on the test tasks of Meta-Dataset in Figure~\ref{fig:attention-score}.

The blue (first head) and orange (second head) heatmaps summarize the values of the attention scores (Equation~\ref{eq:urt-attention}), averaged across several tasks for each test domain. Specifically, the element on row $t$ and column $i$ is the averaged attention scores $\alpha_i$ computed on test set domain $t$ for the backbone from domain $i$. Note that the last two rows are the two unseen domain datasets.
We found that for datasets from the seen domains, i.e.\ the first eight rows, one head (right, orange) consistently puts most of its weight on the backbone pre-trained on the same domain, while the other head (left, blue) learns relatively smoother weight distributions that blends other related domains. 
For unseen datasets, the right head puts half of its weight on ImageNet and the left head learned to blend the representations from four backbones.\looseness=-1

\subsection{URT using FiLM Modulated Backbones}

\begin{wraptable}{r}{0.46\textwidth}
\vspace{-5mm}
\centering
\setlength{\tabcolsep}{2.5pt}
\caption{
Performance comparison using parametric network family (pf) backbones.}\label{table:film-layer}
\begin{tabular}{lcccccc|c}
\toprule
   & \textbf{SUR-pf}~\cite{dvornik2020selecting} & \textbf{URT-pf} & \textbf{VS.} \\
\midrule
ILSVRC &  56.4 $\pm$ 1.2 &  55.5 $\pm$ 1.1 & =  \\ 
Omniglot &  88.5 $\pm$ 0.8 & \textbf{90.2 $\pm$ 0.6} & +\\ 
Aircraft &   79.5 $\pm$ 0.8  & 79.8 $\pm$ 0.7 & = \\
Birds & 76.4 $\pm$ 0.9 & 77.5 $\pm$ 0.8 & = \\ 
Textures & 73.1 $\pm$ 0.7  & 73.5 $\pm$ 0.7& = \\
Quick Draw &  75.7 $\pm$ 0.7 & 75.8 $\pm$ 0.7 & =\\
Fungi & 48.2 $\pm$ 0.9 & 48.1 $\pm$ 0.9 & = \\
VGG Flower &  90.6 $\pm$ 0.5 & \textbf{91.9 $\pm$ 0.5} & + \\
\midrule
Traffic Signs & 65.1 $\pm$ 0.8 & \textbf{67.5 $\pm$ 0.8} & + \\
MSCOCO &  52.1 $\pm$ 1.0 & 52.1 $\pm$ 1.0 & = \\
MNIST & 93.2 $\pm$ 0.4  & 93.9 $\pm$ 0.4 & = \\
CIFAR10   & 66.4 $\pm$ 0.8 & 66.1 $\pm$ 0.8 & = \\
CIFAR100 & 57.1 $\pm$ 1.0  & 57.3 $\pm$ 1.0 & = \\
\bottomrule
\end{tabular}
\vspace{-4mm}
\end{wraptable}
As additional evidence of the benefit of URT on universal representations, we also present experiments based on a different set of backbone architectures. 
Following SUR~\cite{dvornik2020selecting}, we consider the backbones from a parametric network family, obtained by training a base backbone on one dataset (ILSVRC) and then learning separate FiLM layers~\cite{film2018} for each other dataset, to modulate the backbone so it is adapted to the other domains. These backbones collectively have only 0.5\% more parameters than a single backbone.\looseness=-1

A comparison between SUR and URT using these backbones (referred to as SUR-pf and URT-pf) is presented in Table~\ref{table:film-layer}. Once again, URT improves the performance on three datasets without sacrificing performance on others. 
Additionally, URT-pf now achieves better performance than BOHB-E on VGGFlower.

\subsection{Hyper-Parameter and Ablation Studies}
\label{sec:hyper-parameter-selection}

We analyze the importance of the various components of URT's attention mechanism structure and training strategy in Table~\ref{table:ablation}.
First we analyze the importance of using the support set to model queries and/or keys. 
To this end, we consider setting the matrices ${\bf W}^{q}$ / ${\bf W}^{k}$ of the query / key linear transformation to 0, which only leaves the bias term.
We found that the support set representation is most crucial for building the keys (row ${\rm w/o} ~{\bf W}^k$ in the table) and has minor benefits for queries (row ${\rm w/o}~{\bf W}^q$) in the table.
This observation is possibly related to the success of attention-based models with learnable constant queries~\cite{liu2016learning,lin2017structured}.
We also found that adding a regularizer $\Omega(\Theta)$ as in Equation~\ref{eq:regularizer}  is important for some datasets, specifically VGG Flower and Birds.

\begin{table}[h!]
\centering
\setlength{\tabcolsep}{1.2pt}
\caption{
Meta-Dataset performance variation on ablations of elements of the URT layer.
}
\vspace{0.5em}
\begin{tabular}{lccccccccccccc}
\toprule
   & \textbf{\rot{ILSVRC}}  & \textbf{\rot{Omniglot}} & \textbf{\rot{Aircraft}} & \textbf{\rot{Birds}} & \textbf{\rot{Textures}} & \textbf{\rot{Draw}} & \textbf{\rot{Fungi}} & \textbf{\rot{Flower}} & \textbf{\rot{Signs}} &
   \textbf{\rot{MSCOCO}}  \\
\midrule
 ${\rm w/o}~{\bf W}^{q}$ &  +0.2 & -0.2 & -0.6 & -0.1 & -0.3 & -0.2 & 0.0 & -0.2 & -0.8 & -0.1  \\
 ${\rm w/o}~{\bf W}^{k}$ & -14.2 &  -2.8 & -10.7 & -18.1 & -7.6 & -9.3 & -22.4 & -3.6 & -0.26 & -10.9 \\
 ${\rm w/o}~r(S_c)$ & -14.2 & -2.8 & -10.7 & -18.1 & -7.6 & -9.2 & -22.4 & -3.6 & -0.26 & -10.9 \\
 ${\rm w/o}~\Omega(\Theta)$  &  0.0 & -0.9 & -0.4 & -3.3 & -1.2 & -0.2 & +0.3 & -9.0 & -2.0 & 0.0 \\  
\bottomrule
\end{tabular}
\label{table:ablation}
\end{table}

An important hyper-parameter in URT is the number of heads $H$.
We chose this hyper-parameter based on the performance on validation set of tasks in Meta-Dataset. In Table~\ref{table:width-depth}, we show the validation performance of URT for varying number of heads. 
As suggested by \citet{triantafillou2019meta}, we considered looking at the rank of the performance achieved by each choice of $H$ for each validation domains, and taking the average across domains as a validation metric. However, since the performances when using two to four heads are similar and yield the same average rank, we instead simply consider the average accuracy as the selection criteria. 

\begin{table}[h!]
\centering
\setlength{\tabcolsep}{4pt}
\caption{
Validation performance on Meta-Dataset using different number of heads
}
\vspace{0.5em}
\begin{tabular}{lccccccccc}
\toprule
    & \textbf{\rot{1}}  & \textbf{\rot{2}} & \textbf{\rot{3}} & \textbf{\rot{4}} & \textbf{\rot{5}} & \textbf{\rot{6}} & \textbf{\rot{7}} & \textbf{\rot{8}}
    \\
\midrule
Average Accuracy & 74.605 & 77.145 &  76.943 & 76.984 &  76.602 & 75.906 & 75.454  & 74.473 \\
Average Rank & 2.875 & 1.000 & 1.000 & 1.000 & 2.250 & 2.250 & 2.25 & 2.50 \\
\bottomrule
\end{tabular}
\label{table:width-depth}
\end{table}

In general, we observe a large jump in performance when using multiple heads instead of just one.
However, since the number of heads controls the capacity, predictably we also observe that having too many heads leads to overfitting.

\section{Conclusion}
We proposed the URT layer to effectively integrate representations from multiple domains and demonstrated improved performance in multi-domain few-shot classification. 
Notably, our URT approach was able to set a new state-of-the-art on Meta-Dataset, and never performs worse than its predecessor (SUR) while also being $10{\times}$ more efficient at inference.  
This work suggests that combining meta-learning with pre-trained universal representations is a promising direction for new few-shot learning methods. Specifically, we hope that future work can investigate the design of richer forms of universal representations that go beyond simply pre-training a single backbone for each domain, and developing meta-learners adapted to those settings.

\section*{Broader Impact}
Our URT model may present an interesting element of solution for applications that present difficulties in the collection and sharing of data. 
This could include settings where each user of an application has limited private data, and as such desires that a classification task be executed directly and solely on their devices.
Any deployment of the proposed model however should be preceded by an analysis of the potential biases captured by the dataset sources used for training and the correction of any such undesirable biases captured by the pre-trained backbones and model.

\section*{Acknowledgement}
We would like to thank Tianyi Zhou for paper review and suggestions. The computation support for this project is provided by Compute Canada and Google Cloud.
This project was supported by the Canada CIFAR AI Chairs program.

\bibliographystyle{plainnat}{
\bibliography{abrv,neurips_2020}
}

\end{document}